\documentclass[10pt,journal,compsoc]{IEEEtran}

\ifCLASSINFOpdf
  \usepackage[pdftex]{graphicx}
  \DeclareGraphicsExtensions{.pdf,.jpeg,.png}
\else
\fi
\usepackage[cmex10]{amsmath}
\usepackage{bm}
\usepackage{array}
\usepackage{amssymb}

\usepackage {diagbox} 

\usepackage[ruled, lined, linesnumbered]{algorithm2e}

\usepackage{tikz}

\definecolor{taskColor}{rgb}{0.94, 0.87, 0.74}

\tikzstyle{user2Shaded}=[circle, draw, fill=gray!40, inner sep=0pt, minimum size=5mm]
\tikzstyle{task2Shaded}=[rectangle, draw, fill=taskColor, inner sep=0pt, minimum size=5mm]

\usepackage{array}

 \usepackage[caption=false,font=footnotesize]{subfig}
\begin{document}
%
\title{Assessing Four Neural Networks on Handwritten Digit Recognition Dataset (MNIST)}
%
%
%
%

\author{Feiyang Chen, Nan Chen, Hanyang Mao, Hanlin Hu}

%
%

\markboth{Chuangxinban Journal of computing, June~2018}%
{Shell \MakeLowercase{\textit{et al.}}: Bare Demo of IEEEtran.cls for Computer Society Journals}
%



\IEEEtitleabstractindextext{%
\begin{abstract}
Although the image recognition has been a research topic for many years, many researchers still have a keen interest in it[1]. In some papers[2][3][4], however, there is a tendency to compare models only on one or two datasets, either because of time restraints or because the model is tailored to a specific task. Accordingly, it is hard to understand how well a certain model generalizes across image recognition field[6]. In this paper, we compare four neural networks on MNIST dataset[5] with different division. Among of them, three are Convolutional Neural Networks (CNN)[7], Deep Residual Network (ResNet)[2] and Dense Convolutional Network (DenseNet)[3] respectively, and the other is our improvement on CNN baseline through introducing Capsule Network (CapsNet)[1] to image recognition area. We show that the previous models despite do a quite good job in this area, our retrofitting can be applied to get a better performance.  The result obtained by CapsNet is an accuracy rate of 99.75\%, and it is the best result published so far. Another inspiring result is that CapsNet only needs a small amount of data to get the excellent performance. Finally, we will apply CapsNet's ability to generalize in other image recognition field in the future.
\end{abstract}

\begin{IEEEkeywords}
Neural Network, CNN, CapsNet, DenseNet, ResNet, MNIST.
\end{IEEEkeywords}}

\maketitle

\IEEEdisplaynontitleabstractindextext

%
\IEEEpeerreviewmaketitle

\IEEEraisesectionheading{\section{Introduction}\label{sec:introduction}}
Motivated by the development of artificial intelligence, there has been a good amount of progress in image recognition over the past 10 years, including the proposal of many new models and the creation of benchmark datasets. 
\par
In some papers, however, there is a tendency to compare models only on one or two datasets, either because of time restraints or because the model is tailored to a specific task. Accordingly, it is hard to understand how well a certain model generalizes across image recognition field. 
\par
In this paper, our main contributions are, therefore, comparing four mainstream image recognition models on MNIST dataset with different division. Among of them, three are CNN, ResNet and DenseNet respectively. These three models are already proved to have good performance in image recognition, and we summarize the characteristics of these models in Section 2. In addition, we find that the standard CNN model still exists some drawbacks. Accordingly we use CNN as a baseline model, and improve it through applying CapsNet to optimize on this basis. It is the fourth model and is described in detail in Section 3. We use the MNIST dataset for the test because the recognition of handwritten digits is a topic of practical importance. This object has continued to produce much research effort in recent years for several reasons. First, standard benchmark datasets like MNIST exist that make it easy for us to obtain result. Second, many publications and techniques are available that can be cited and built on, respectively. In order to make the model more generalizable in the field of image recognition, we randomly divided the MNIST dataset into 25\%, 50\%, 75\%, and 100\% to test. 
\par
Ultimately, we contribute to a better understanding of the performance of different model architectures on MNIST dataset. Consequently, we detect that CapsNet is the best overall model, which outperforms the other models on all tasks and consistently beats the baseline. The results of experiment are presented in Section 4, and conclusion follows in Section 5.

\section{Related works}
This section describes MNIST dataset which will be used in the experiments and
then discusses the characteristics of the three neural network models.
 \subsection{Dataset}
\begin{figure}[ht]
 	
 	\centering
 	\includegraphics[scale=0.4]{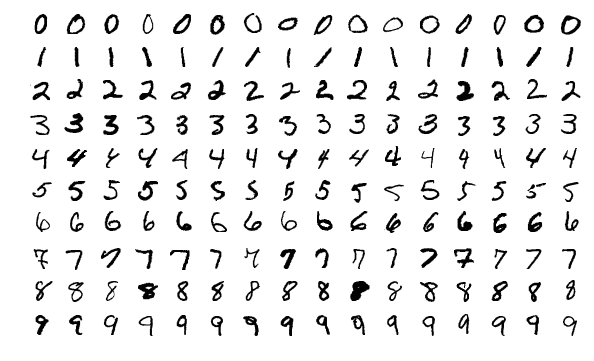}
 	
 	\centering\caption{Example images from the MNIST dataset, including 60,000 images in the training set and 10,000 patterns in the testing set. }
 	\label{fig:mn}
 \end{figure}

 The MNIST dataset is from the National Institute of Standards and Technology (NIST).
 The training set consists of handwritten numbers from 250 different people, of which 50\% are high school students and 50\% are from the Census Bureau. The test set is also the same proportion of handwritten digital data. MNIST dataset totally contains 60,000 images in the training set and 10,000 patterns in the testing set, each of size 28×28 pixels with 256 gray levels[8]. The dataset can be downloaded online and some examples from the MNIST corpus are shown in Fig. \ref{fig:mn}.
 
\subsection{CNN}

In machine learning, CNN is a feed-forward artificial neural networks, most commonly applied to analyzing visual imagery. For CNN, the earliest date can be traced back to the 1986 BP algorithm[9]. Then in 1989 LeCun used it in multi-layer neural networks[10]. Until 1998, LeCun proposed the LeNet-5 model, and the neural network prototype was completed. CNN consists of one or more
convolutional layers and the top fully connected layer, and it also includes associated weights and a pooling layer. This structure allows the convolutional neural network to take advantage of the two dimensional structure of the input data, so it can give very good results in image recognition[11]. So we try to apply it to the MNIST dataset for testing.

\subsection{ResNet}
\begin{figure}[ht]
	
	\centering
	\includegraphics[scale=0.53]{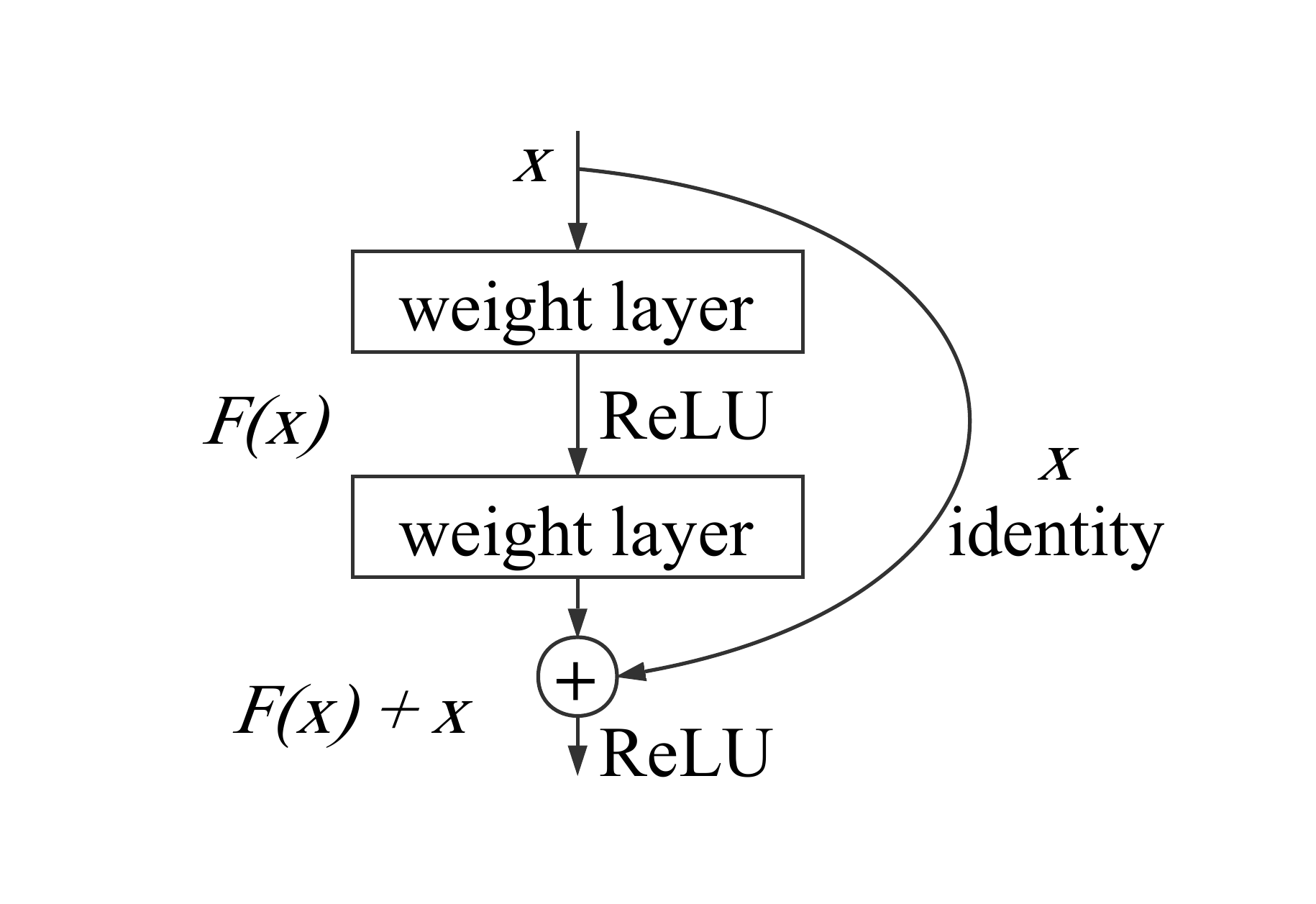}
	
	\centering\caption{ Unit module block, where x means the input and F(x) means the output of the weight layer, the final output is the sum of F(x) and x. }
	\label{fig:unit}
\end{figure}

Deep convolutional neural networks have led to a series of breakthroughs for image classification. However, when deeper networks are able to start converging, a degradation problem[12] has been exposed: with the network depth increasing, accuracy gets saturated and then degrades rapidly. Therefore, ResNet is presented in 2017. It can reduce the train\_error while deepening the depth of the network, and solve the problem of gradient dispersion[13], improving network performance, which is shown in the Eq. \ref{res}. Most importantly, ResNet can not only be very deep, but also has a very simple structure. It is a very small single module piled up, its unit module block as shown in Fig. \ref{fig:unit}.
\begin{equation}\label{res}
x_{l} = H_{l}(x_{l-1}) + x_{l-1}
\end{equation}
In the Eq. \ref{res}, $l$ represents layer, $x_{l}$ represents the output of the $l$ layer, $H_{l}$ represents a nonlinear transformation. For ResNet, the output of the $l$ layer is the output of the $l-1$ layer plus the nonlinear transformation of the output of the $l-1$ layer.
\par
So we try to apply it to the MNIST dataset for testing.
\subsection{DenseNet}
In the field of image recognition, CNN has become the most popular method. A milestone in the history of CNN is the emergence of the ResNet model[14]. ResNet can train deeper CNN models to achieve higher accuracy. The basic idea of the DenseNet model is the same as that of ResNet, but it establishes a dense connection between all previous and subsequent layers[15]. Its other major feature is feature reuse through the connection of features on the channel. Therefore we also tested its performance on the MNIST dataset.



\section{Experimental setup}
We compare four models, three of which are mentioned in Section 2. The other is our retrofitting and improvement based on CNN model. It is described in detail in Section 4.2 and 4.3. We use the MNIST datasets mentioned in Section 2.1 to test these models. 
\par
We tested all the models using a workstation built from commodity hardware: dual GeForce GTX 1080 graphics cards, an i7-6800K CPU, and 64 GB of RAM. Our implementation is in TensorFlow and we use the Adam optimizer with TensorFlow default parameters, including the exponentially decaying learning rate, to minimize the sum of the margin losses.

\subsection{Baseline}
Our model is based on a standard CNN with three convolutional layers, which is demonstrated to achieve a low test error rate on MNIST. The channels of three layers are 256, 256, 128 respectively. Each layer has 5 $\times$ 5 kernels and stride of one. Followed by the last convolutional layers are two fully connected layers of size 328, 192. After that is a 10 class softmax with cross entropy loss.
However, the baseline model has two shortcomings. First, training a powerful CNN model requires a large number of training data. Second, in the pooling layer, CNN loses some of the information, which leads to the ignorance of interrelationships between different component[16].
\par
Thus, for small changes in input, the output of CNN will be almost constant, which may result in a higher error rate.

\subsection{Retrofitting}

In order to overcome these shortcomings of CNN, we try to introduce CapsNet to optimize the baseline. Fig. \ref{fig:cap} shows the structure of the CapsNet. CapsNet uses capsules instead of neurons. The input and output of the capsule are high-dimensional vectors, where the module length represents the probability of occurrence of an object, and the direction represents the position, color, size and other information. The output of the low-level capsules is used to generate a prediction through transformation matrices, which are then linearly integrated and passed into high-level capsules according to certain weights. The method of updating the weights is a dynamic routing algorithm, which compares the output of high-level capsules with the prediction of low-level capsules, and increases the input weights of low-level capsules with higher similarity until convergence.
\par
Through the capsule, we retain the information on the details of the picture. In this way, on the basis of accurately identifying the image, small changes in the image input will cause small changes in the output. It has a human-like understanding of the three-dimensional space. In addition, with less information loss, it only needs a small amount of data to achieve amazing results compared to CNN.
\def\layersep{2cm}
\begin{figure}[ht]
	\centering
	\begin{tikzpicture}[shorten >=1pt,->,draw=black!50, node distance=\layersep]
	\tikzstyle{every pin edge}=[<-,shorten <=1pt]
	\tikzstyle{neuron}=[circle,fill=black!25,minimum size=17pt,inner sep=0pt]
	\tikzstyle{input neuron}=[neuron, fill=green!50];
	\tikzstyle{output neuron}=[neuron, fill=red!50];
	\tikzstyle{hidden neuron}=[neuron, fill=blue!50];
	\tikzstyle{annot} = [text width=4em, text centered]
	
	\foreach \name / \y in {1,...,2}
	\node[input neuron] (I-\name) at (0,-\y*2.5) {$u_{\y}$};
	
	\foreach \name / \y in {1,...,8}
	\path[yshift=0.5cm]
	node[hidden neuron] (H-1) at (\layersep,-1 cm) {$\hat{u}_{1|1}$};
	\path[yshift=0.5cm]
	node[hidden neuron] (H-2) at (\layersep,-2 cm) {$\hat{u}_{1|2}$};
	\path[yshift=0.5cm]
	node[hidden neuron] (H-3) at (\layersep,-3 cm) {$\hat{u}_{2|1}$};
	\path[yshift=0.5cm]
	node[hidden neuron] (H-4) at (\layersep,-4 cm) {$\hat{u}_{2|2}$};
	\path[yshift=0.5cm]
	node[hidden neuron] (H-5) at (\layersep,-5 cm) {$\hat{u}_{3|1}$};
	\path[yshift=0.5cm]
	node[hidden neuron] (H-6) at (\layersep,-6 cm) {$\hat{u}_{3|2}$};
	\path[yshift=0.5cm]
	node[hidden neuron] (H-7) at (\layersep,-7 cm) {$\hat{u}_{4|1}$};
	\path[yshift=0.5cm]
	node[hidden neuron] (H-8) at (\layersep,-8 cm) {$\hat{u}_{4|2}$};
	\foreach \name / \y in {1,...,4}
	\path[yshift=0.5cm]
	node[output neuron,pin={[pin edge={->}]right:Squashing}] (O-\name) at (\layersep*2,-\y*1.77 cm) {$S_{\y}$};
	
	\path (I-1) edge node[above]  {\small $W_{11}$}(H-1);
	\path (I-1) edge node[above]  {\small $W_{12}$}(H-3);
	\path (I-1) edge node[above]  {\small $W_{13}$}(H-5);
	\path (I-1) edge node[below]  {\small $W_{14}$}(H-7);
	
	\path (I-2) edge node[below]  {\small $W_{21}$}(H-2);
	\path (I-2) edge node[above]  {\small $W_{22}$}(H-4);
	\path (I-2) edge node[above]  {\small $W_{23}$}(H-6);
	\path (I-2) edge node[above]  {\small $W_{24}$}(H-8);

	\path (H-1) edge node[above]  {\small $C_{11}$}(O-1);
	\path (H-2) edge node[above]  {\small $C_{21}$}(O-1);
	
	\path (H-3) edge node[above]  {\small $C_{12}$}(O-2);
	\path (H-4) edge node[above]  {\small $C_{22}$}(O-2);
	
	\path (H-5) edge node[above]  {\small $C_{13}$}(O-3);
	\path (H-6) edge node[above]  {\small $C_{23}$}(O-3);
	
	\path (H-7) edge node[above]  {\small $C_{14}$}(O-4);
	\path (H-8) edge node[above]  {\small $C_{24}$}(O-4);
	
	\node[annot,above of=H-1, node distance=0.5cm] (hl) {$\hat{u}_{j|i}$};
	\node[annot,left of=hl, node distance=1.1cm](h2) {$W_{ij}$};
	\node[annot,left of=h2, node distance=1cm](h3) {$u_{i}$};
	\node[annot,right of=hl, node distance=1.0cm](h4) {$C_{ij}$};
	\node[annot,right of=h4, node distance=1.0cm](h5) {$S_{j}$};
	\end{tikzpicture}
	\caption{Structure of CapsNet, where $u_{i}$ is the input layer, $W_{ij}$ is the weight matrix, $\hat{u}_{j|i}$ is the $u_{i}$'s prediction to $S_{j}$, $C_{ij}$ is the weight and $S_{j}$ is the output layer. Squashing is the activate function as shown in Eq. \ref{eq}.}
	\label{fig:cap}
\end{figure}
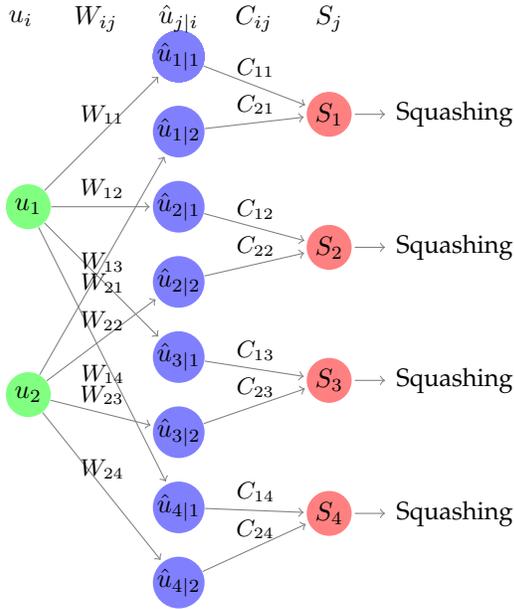
\begin{equation}\label{eq}
v_{j} = \frac{||s_{j}||^{2}}{1+||s_{j}||^{2}}\frac{s_{j}}{||s_{j}||}
\end{equation}
\subsection{CapsNet Architecture}
\begin{figure}[ht]
	
	\centering
	\includegraphics[scale=0.26]{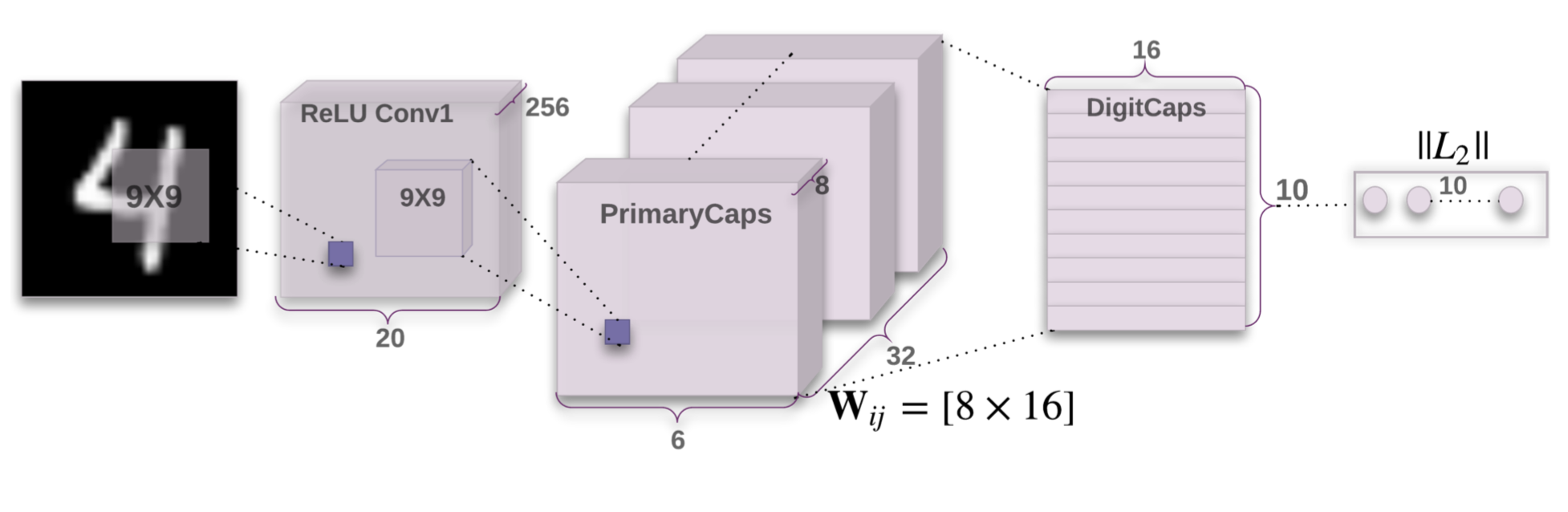}
	
	\centering\caption{A simple CapsNet with three layers. The convolutional layer extracts image features, PrimaryCaps integration the features, and DigitCaps output the prediction.}
	\label{fig:arc}
\end{figure}

The architecture is showed in Fig. \ref{fig:arc}, it consists of one convolutional layer and two capsule layers[18][19]. The convolutional layer 1 has 256, 9 $\times$ 9 convolution kernels with a stride of 1 and ReLU activation. This layer extracts the basic features of the image, and then uses them as the inputs of the primary capsules layer (PrimaryCaps). The PrimaryCaps contains 32 capsules, which receives the basic features detected by the convolution layer, creating a combination of features. The 32 primary capsules in this layer are essentially similar to the convolutional layer[20]. Each has 8, 9 $\times$ 9 $\times$ 256 convolution kernels with a stride of 2. The output of PrimaryCaps is 6632 eight-dimensional vector. The last layer is digital capsules  layer (DigitCaps), it has 10 digital capsules and each of which represents the prediction of number. Every capsule receives input from all capsules in the PrimaryCaps, and finally outputs the result.
\section{Result}
\begin{figure}[ht]
	
	\centering
	\includegraphics[scale=0.57]{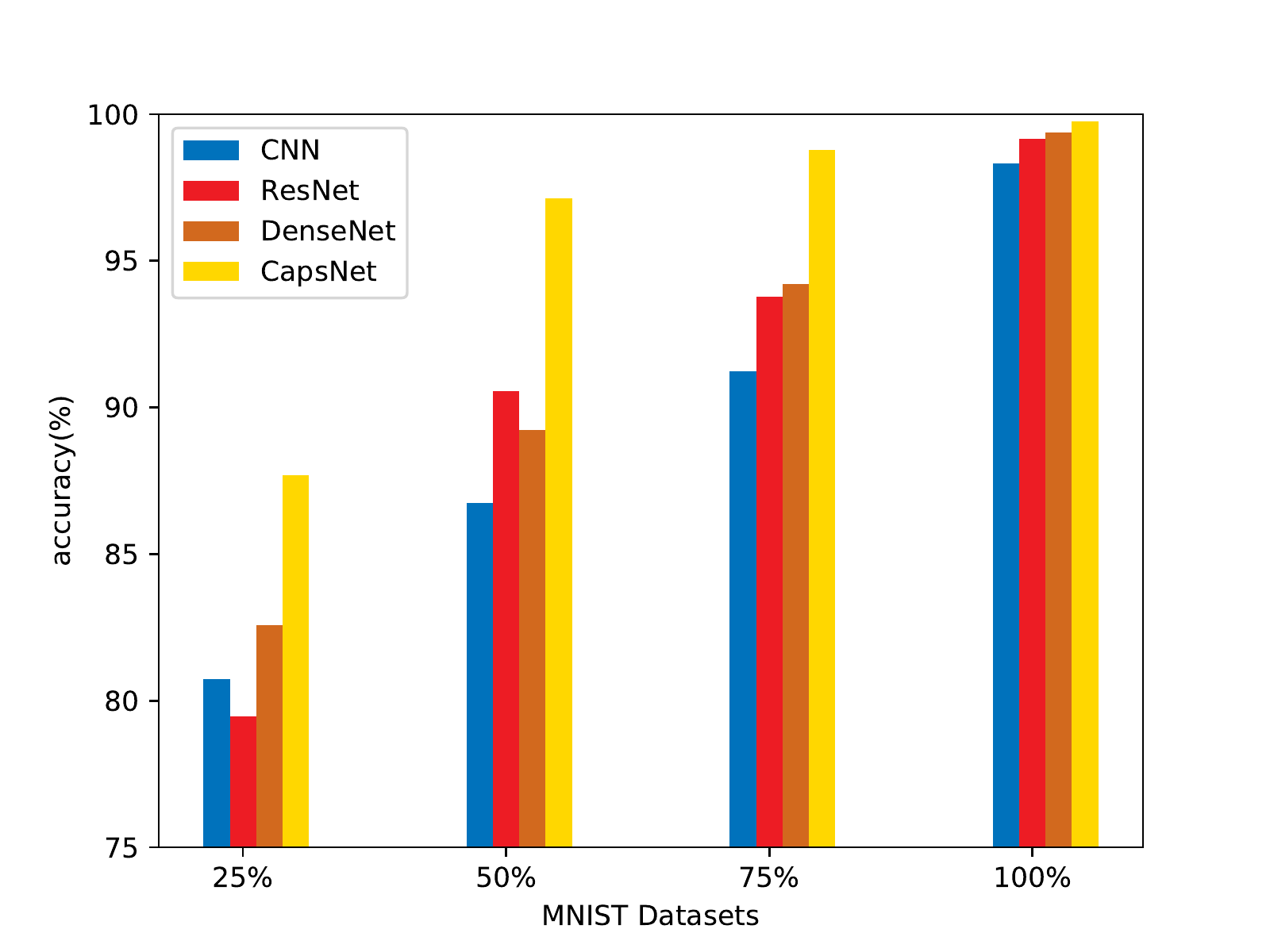}
	
	\centering\caption{ The results for the four models across all divided datasets. }
	\label{fig:test}
\end{figure}

We randomly divided the MNIST dataset into 25\%, 50\%, 75\%, and 100\%. Table \ref{my-label} shows the results for the four models across all divided datasets, and we visualize them in Fig. \ref{fig:test}. Obviously, CNN continues to be a strong baseline: Though it never provides the best result on a dataset, it gives better results than ResNet on 25\% MNIST. Because the ResNet's network structure requires a larger number of data to train. DenseNet performs better than CNN on all divided datasets. It also improves the results of ResNet across all datasets but 50\% dataset. That is related to DenseNet's parameter settings. Inspiringly, CapsNet is the best overall model, which outperforms the other models on all tasks and consistently beats the baseline. In addition, we can observe from the Table \ref{my-label} that CapsNet trained with half datasets reach approximately equal accuracy with complete CNN. We attribute this to CapsNet's ability to generalize in image recognition. This is in line with other research[14][16][17], which suggests that this model is very robust across tasks as well as datasets.
\renewcommand\arraystretch{1.51}
\begin{table}[ht]
	\centering
	\caption{Results of experiment on divided datasets.}
	\label{my-label}
	\begin{tabular}{|l|c|c|c|c|}
		\hline
		\diagbox{Models}{Accuracy(\%)}{MNIST}
		& 25\%  & 50\%  & 75\%  & 100\% \\
		\hline
		CNN      & 80.73 & 86.73 & 91.23 & 98.32 \\
		\hline
		ResNet   & 79.46 & 90.55 & 93.78 & 99.16 \\
		\hline
		DenseNet & 82.57 & 89.24 & 94.20 & 99.37 \\
		\hline
		CapsNet  & 87.68 & 97.12 & 98.79 & 99.75\\
		\hline
	\end{tabular}

\end{table}

\section{Conclusion}\label{sec:conclusion}

The goal of this paper has been to discover which models perform better across divided MNIST datasets. We compared four models on MNIST dataset with different division, and showed that CapsNet perform best across datasets. Additionally, we also observe surprisingly that CapsNet requires only a small amount of data to achieve excellent performance. Finally, we will apply CapsNet's ability to generalize in other image recognition field in the future.


%

\ifCLASSOPTIONcompsoc
  \section*{Acknowledgments}
\else
  \section*{Acknowledgment}
\fi

The authors would like to thank Guodong Sun, Yanyan Xu and Dengfeng Ke for their invaluable technical support and constructive discussion of the ideas during this research.

\ifCLASSOPTIONcaptionsoff
  \newpage
\fi



%

%



\begin{IEEEbiography}
	[{\includegraphics[width=1in,height=1.25in,clip,keepaspectratio]{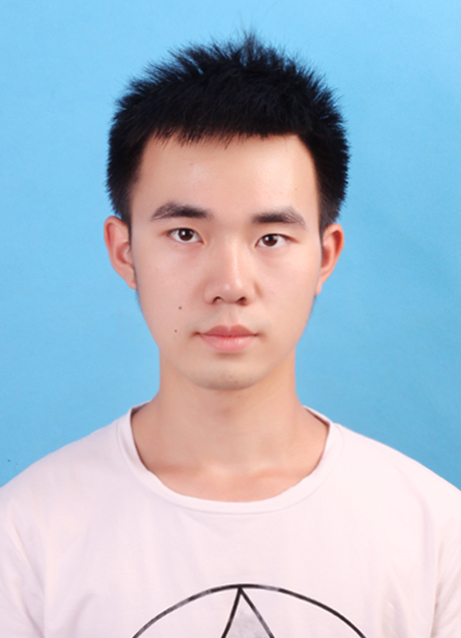}}]{Feiyang Chen}
	was born on October 29th, 1998, in Hunan, China. He got admitted by Beijing Forestry University and majored Computer Science in 2016. Where he received multiple awards, including Honorable Mention in Mathematical Contest In Modeling (MCM), National University Student Mathematics Competition Third Prize, "triple-A student" of the university, "Excellent student cadre", "Honored Student" of the year and so on.
	\par
	From May, 2018 till now, he was an researcher at the AI lab of School of CS, BJFU. Where he hosts an collage innovation training program:"Research on Text Sentiment Analysis Algorithm Based on CNN and LSTM". His research interests include Natural Language Processing, Machine Learning, Deep Reinforcement Learning and Text Sentiment Analysis.
\end{IEEEbiography}

\begin{IEEEbiography}
	[{\includegraphics[width=1in,height=1.25in,clip,keepaspectratio]{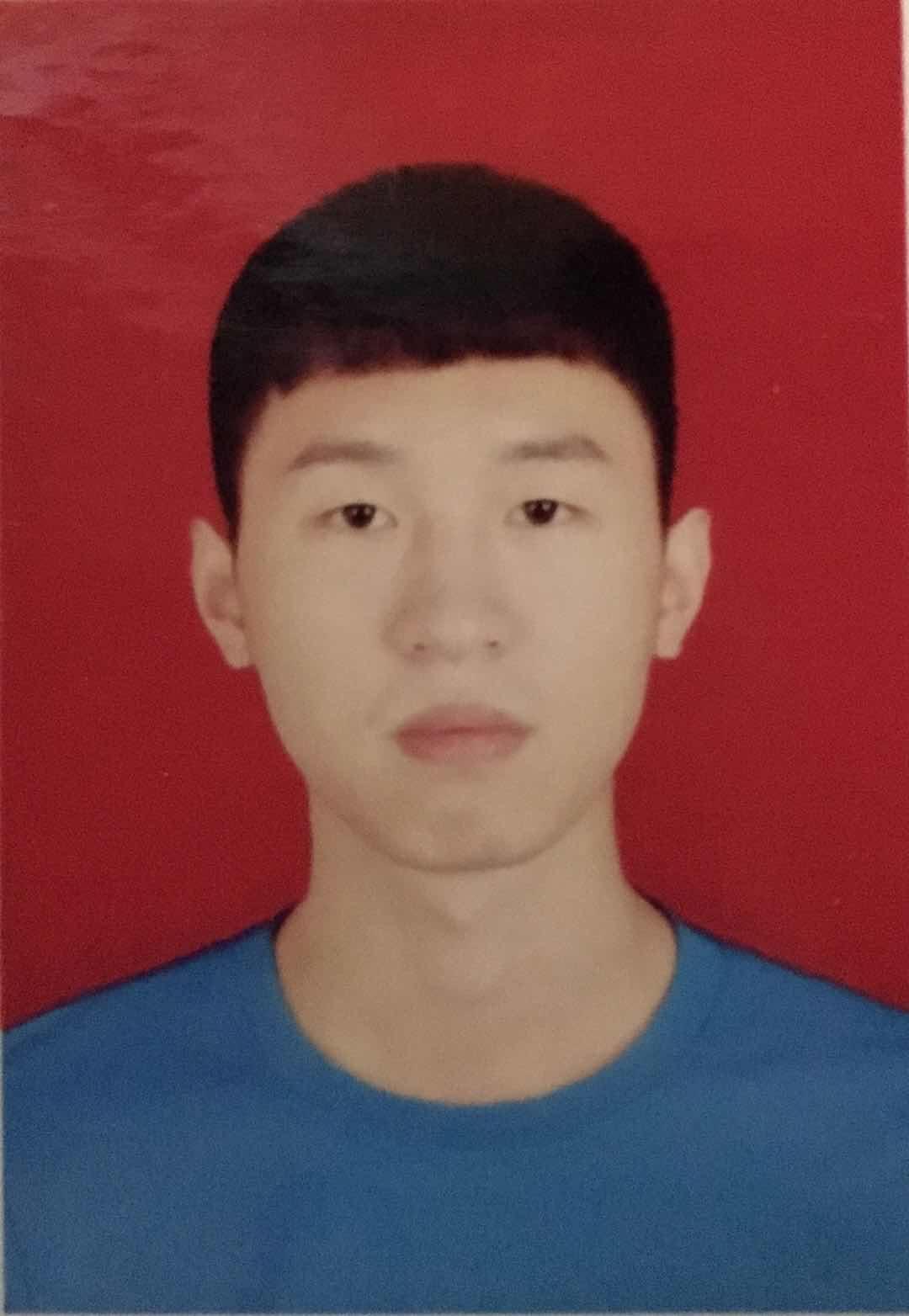}}]{Nan Chen}
	 was born on September 24th, 1997, in Beijing, China. He got admitted by Beijing Forestry University and majored Computer Science in 2016. Where he received multiple awards, including "Third prize" in laoqiao - competiton, "triple-A student" of the university, "Excellent student cadre", "Honored Student" of the year and so on.
	 \par
	During the summer of 2017, he participated in the Collage student social activity contest, with the excellence behavior in developing WeChat Applets, he was awarded "Outstanding Participant" and his product was awarded "Second Prize" in the contest.
	\par
	From May, 2018 till now, he participated in an collage innovation training program:"Research on Text Sentiment Analysis Algorithm Based on CNN and LSTM".
\end{IEEEbiography}

\begin{IEEEbiography}
	[{\includegraphics[width=1in,height=1.25in,clip,keepaspectratio]{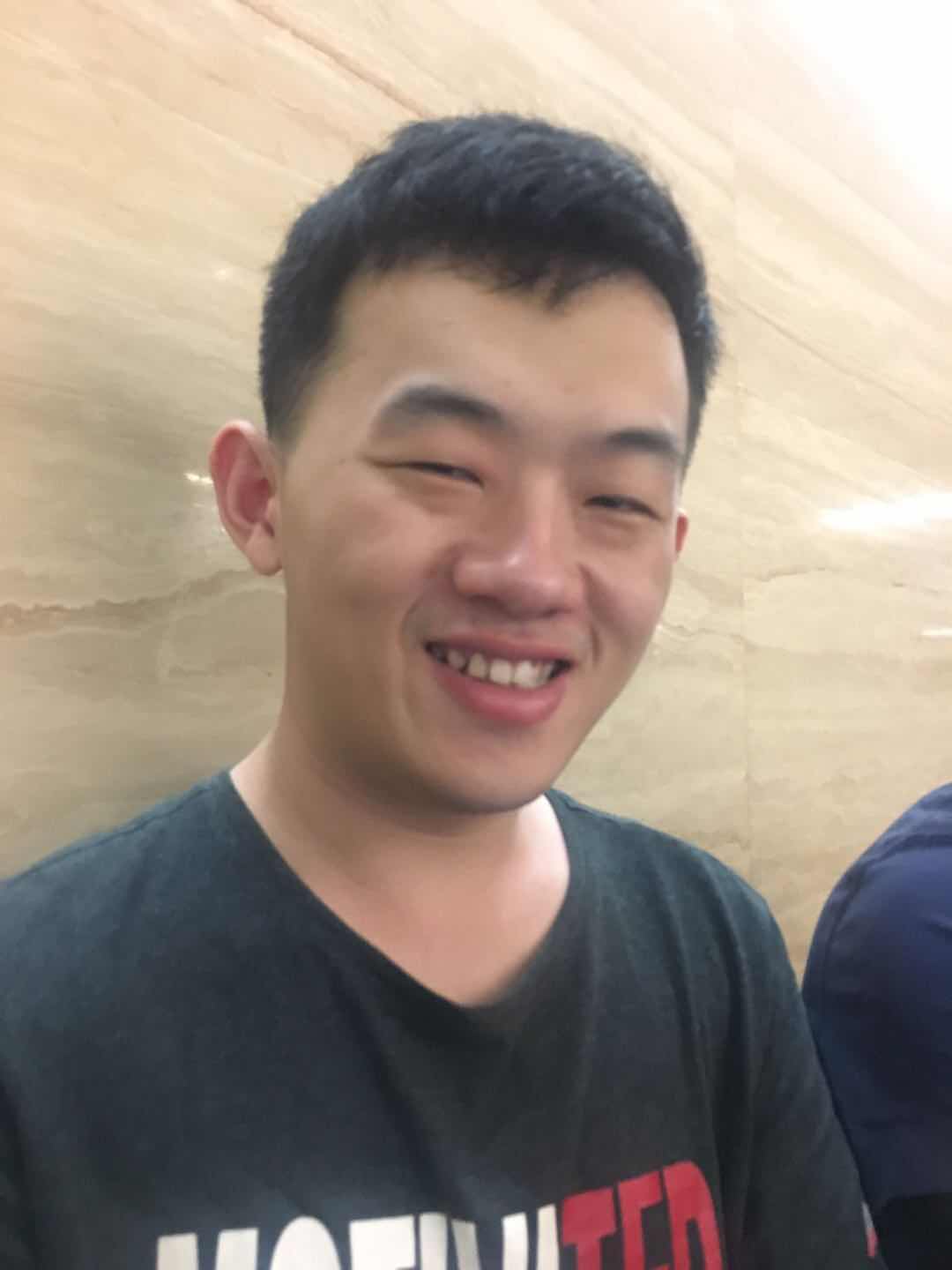}}]{Hanyang Mao}
	  was born in Zhejiang, China in November ,1st 1997. Now he is a student in Beijing Forest University, majoring in computer science. 
	  \par
	 He has a keen interest in algorithm and artificial intelligence. He had won the prize in some algorithm competitions and was a researcher at the AI lab of School of CS, BJFU. He is also good at many kinds of chess, such as Chinese chess, Japanese chess and Gobang. He used to program a Gobang AI by himself. 
	 \par
	 During the summer of 2017, he participated in a group to create WeChat mini program named "Play BJFU", where he hosted the partition of "Course Guide". And in 2018 he joined a program named "Research on the Application of Capsule Network in the Detection of sensitive Words".
\end{IEEEbiography}

\begin{IEEEbiography}
	[{\includegraphics[width=1in,height=1.25in,clip,keepaspectratio]{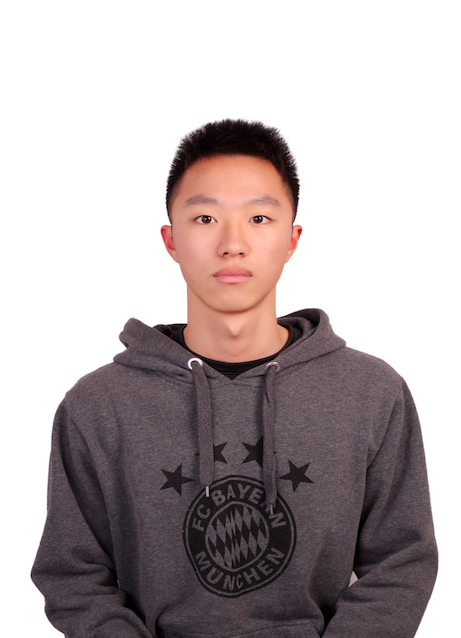}}]{Hanlin Hu}
	 was born in Beijing, China in December, 15th 1997. He got admitted by Beijing Forestry University and majored Computer Science there.  When he was in primary school, he used to study chess and awarded many prize from some competitions. He also has a deep love for sport, especially basketball. He is a member of the basketball team of Information Science and Technology. During the summer of 2018, he participate in the collage innovation training program. And the program he host named "Research on the Application of Capsule Network in the Detection of sensitive Words". From September 2017 till now, he was learning at Artificial intelligence lab of BJFU. At there he got a better understanding of Speech Recognition.
\end{IEEEbiography}




\end{document}